\documentclass[conference]{IEEEtran}
\usepackage{cite}
\usepackage{amsmath,amssymb,amsfonts}
\usepackage{graphicx}
\usepackage{textcomp}
\usepackage{xcolor}
\usepackage{multirow}
\usepackage{algorithm, algpseudocode}
\usepackage{booktabs}

\def\BibTeX{{\rm B\kern-.05em{\sc i\kern-.025em b}\kern-.08em
    T\kern-.1667em\lower.7ex\hbox{E}\kern-.125emX}}

\begin{document}

\title{ Permutation-Invariant Tabular Data Synthesis}

\author{\IEEEauthorblockN{Yujin Zhu}
\IEEEauthorblockA{\textit{Computer Engineering} \\
\textit{TU Delft}\\
Delft, the Netherlands \\
Y.Zhu-17@student.tudelft.nl}
\and
\IEEEauthorblockN{Zilong Zhao\IEEEauthorrefmark{1}\thanks{\IEEEauthorrefmark{1}Contact author.}}
\IEEEauthorblockA{\textit{Computer Science} \\
\textit{TU Delft}\\
Delft, the Netherlands \\
Z.Zhao-8@tudelft.nl}
\and
\IEEEauthorblockN{Robert Birke}
\IEEEauthorblockA{\textit{Computer Science} \\
\textit{University of Torino}\\
Torino, Italy \\
birke@ieee.org}
\and
\IEEEauthorblockN{Lydia Y. Chen}
\IEEEauthorblockA{\textit{Computer Science} \\
\textit{TU Delft}\\
Delft, the Netherlands \\
LydiaYChen@ieee.org}
}
\IEEEoverridecommandlockouts
\IEEEpubid{\makebox[\columnwidth]{2022 IEEE International Conference on Big Data\hfill} 
\hspace{\columnsep}\makebox[\columnwidth]{ }}

\maketitle
\IEEEpubidadjcol
\begin{abstract}
Tabular data synthesis is an emerging approach to circumvent strict regulations on data privacy while discovering knowledge through big data. Although state-of-the-art AI-based tabular data synthesizers, e.g., table-GAN, CTGAN, TVAE, and CTAB-GAN, are effective at generating synthetic tabular data, their training is sensitive to column permutations of input data. In this paper, we first conduct an extensive empirical study to disclose such a property of permutation invariance and an in-depth analysis of the existing synthesizers. We show that changing the input column order {worsens} the statistical difference between real and synthetic data by up to 38.67\% due to the encoding of tabular data and the network architectures. To fully unleash the potential of big synthetic tabular data, we propose two solutions: (i) AE-GAN, a synthesizer that uses an autoencoder network to represent the tabular data and GAN networks to synthesize the latent representation, and (ii) a feature sorting algorithm to find the suitable column order of input data for CNN-based synthesizers.
We evaluate the proposed solutions on five datasets in terms of the sensitivity to the column permutation, the quality of synthetic data, and the utility in downstream analyses. Our results show that we enhance the property of permutation-invariance when training synthesizers and further improve the quality and utility of synthetic data, up to 22\%, compared to the existing synthesizers.

\end{abstract}


\section{Introduction}
As one of the most common data types, tabular data are ubiquitous in the operation of banks, governments, hospitals, and manufacturers, which lay the foundations of modern society~\cite{ryan2020deep}. The synthesis of realistic tabular data, i.e., generating synthetic tabular data that are statistically similar to the original data, is crucial for many applications, such as data augmentation~\cite{chen2019faketables}, imputation~\cite{gondara2018mida, camino2020working}, and re-balancing~\cite{engelmann2021conditional, quintana2019towards, koivu2020synthetic}. Another important application is to use generated data to overcome data sharing restrictions~\cite{tablegan} caused by regulations on data protection and privacy, such as European General Data Protection Regulation (GDRP)~\cite{gdpr}. 

Synthesizing realistic tabular data is a non-trivial task. Compared to image and language data, tabular data are heterogeneous - they contain dense continuous features and sparse categorical features. The former can have multiple modes, whereas the latter often have highly-imbalanced distributions~\cite{DNN_and_Tabular_Data_Survey}. Furthermore, the correlation between features in tabular data is often more complex than the spatial or semantic correlation in image or language data. Related features can be far apart spatially, and multiple features can be inter-correlated.

\begin{figure}[t]
    \centering
    \includegraphics[width=1\linewidth]{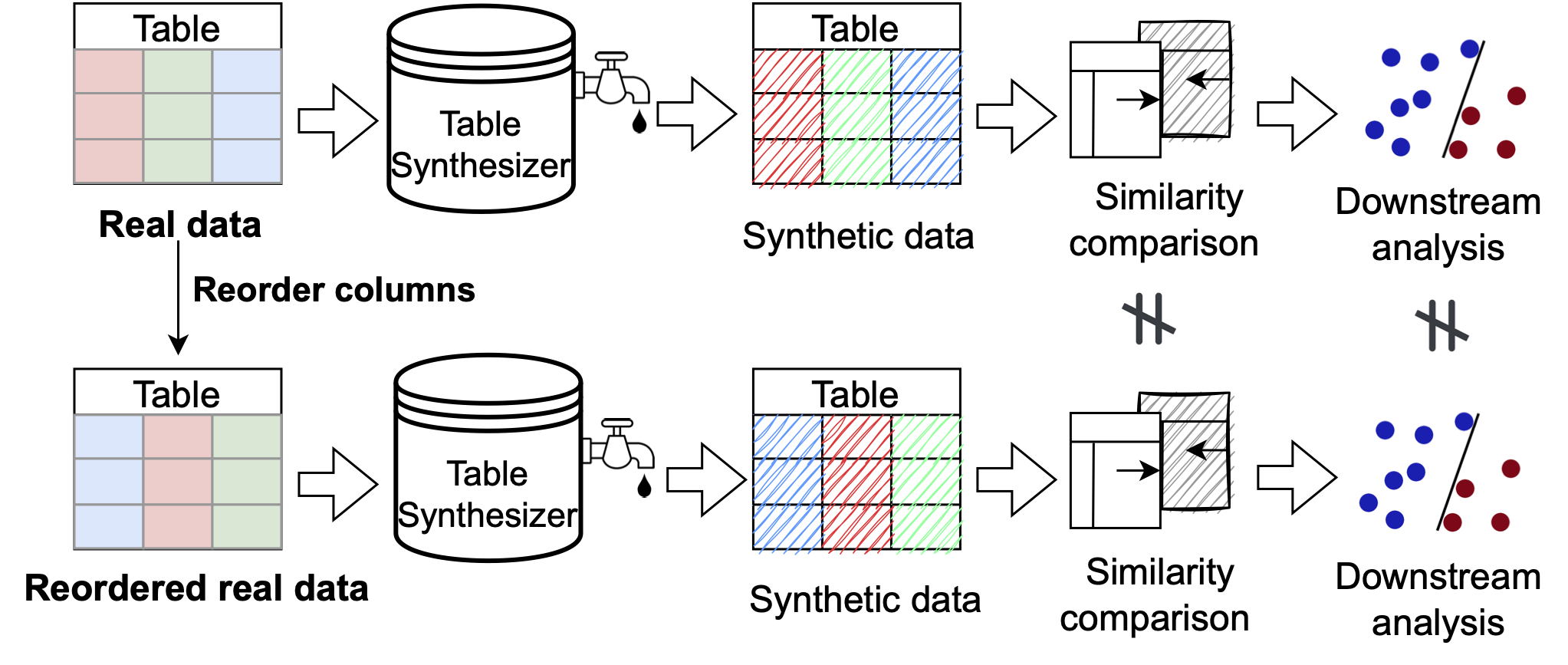}
    \caption{Illustration of lacking column permutation invariance of training tabular data synthesizer. This leads to dissimilarity between real and synthetic data and different downstream analysis results. }
    \label{fig:motivation_compare_similarity}
\end{figure}

Although the state-of-the-art AI-based tabular data synthesizers show promising results~\cite{ctgan&tvae, tablegan, zhao2021ctab, ctabplus}, they suffer from a critical and undiscovered limitation: their training is sensitive to column permutations. Changing the input column order during training influences the quality of the synthetic data, e.g., statistical similarity with real data (see Figure~\ref{fig:motivation_compare_similarity}). Theoretically, reordering the columns of the training input shall not change the capability of synthesizers because the position of columns does not imply any semantic information. We call this property \textit{column permutation invariance}, i.e., the output of generative models is invariant to the column order of their training input. However, our extensive empirical analysis shows that the statistical difference between real and synthetic data increases by up to {38.67\%} after changing the input column order. The main reason is the sparsity issue caused by one-hot data encoding for categorical features and mode-specific normalization for numerical features.

In this paper, we address the limitation of being sensitive to input column permutations and striking a tradeoff between the quality of synthetic data and the training time by two approaches. Our first solution is to leverage an autoencoder~\cite{tschannen2018recent} to improve the representation of tabular data and then design a Wasserstein GAN with gradient penalty (WGAN-GP)~\cite{gulrajani2017improved} based on fully connected networks to synthesize the latent representation. This solution is named AE-GAN. Our second solution is to leverage a feature sorting algorithm that is capable of exploring the correlation among columns and provides the permutation that enhances the synthesizer training, especially for the ones based on convolutional neural networks.


We evaluate both solutions on five real-world machine learning datasets against four state-of-the-art tabular data synthesizers: table-GAN~\cite{tablegan}, CTGAN~\cite{ctgan&tvae}, CTAB-GAN~\cite{zhao2021ctab}, and TVAE~\cite{ctgan&tvae}. The results show that, compared to the baselines, AE-GAN makes a better trade-off among more permutation-invariance, generating high-quality data, and leading to accurate downstream analyses. 
Moreover, its training time is significantly shorter than CTAB-GAN, the best performing model in synthesizing realistic tabular data. Besides, after applying the feature sorting algorithm to CTAB-GAN, the statistical difference between real and synthetic data is improved by 22\% averaged across five datasets.

The contributions of this paper are as follows: 
\begin{itemize}
\item The first empirical study to analyze sensitivity to column permutations for training tabular data synthesizer which reveals its root cause, i.e., data representation and sparsity.

\item A novel tabular data synthesizer, AE-GAN, which effectively achieves high permutation invariance and good quality of synthetic data in terms of statistical similarity and machine learning utility.

\item A feature sorting algorithm for tabular data synthesizers, which helps preserve the relation between highly-correlated features.
\end{itemize}

\section{Background and Related Work}
In this section, we first provide the background on two methods, GAN and autoencoder, which are the cores of the prior art. Then, we compare the related studies in terms of their network structures and data encoding schemes. 

\subsection{Generative Adversarial Network}
Generative adversarial networks (GAN) are a recently developed algorithm~\cite{goodfellow2014generative} for synthetic data generation. 
A GAN consists of two components: a generator ($G$) that learns to produce realistic synthetic data, and a discriminator ($D$) that tries to distinguish real data from synthetic (fake) data. Both $G$ and $D$ are neural networks, e.g., Fully-Connected Networks (FCNs) or Convolutional Neural Networks (CNNs). In the training process, $G$ and $D$ play an adversarial game described as follows:
\begin{equation}
\begin{split}
    \underset{G}{min}\ \underset{D}{max}\ V(G,D) = & \mathbb{E}[log D(x)]_{x \sim p_{data}(x)} \\
    & + \mathbb{E}[log(1-D(G(z)))]_{z \sim p(z)},
\end{split}
\end{equation}
where $x$ is the real sample, $z$ is the random input signal given to $G$, $G(z)$ is the synthetic sample, and $D(\cdot)$ is the probability of a sample being real from the perspective of $D$. The goal of $G$ is to minimize the chance that its generated samples are identified as synthetic, whereas $D$ maximizes the chance of correctly distinguishing real and synthetic samples.

\subsection{Autoencoder}
An autoencoder (AE) is an unsupervised learning algorithm that learns a mapping from high-dimensional inputs to low-dimensional representations~\cite{tschannen2018recent, ng2011sparse}, namely latent vectors. It consists of two models, an encoder ($Enc$) and a decoder ($Dec$). $Enc$ takes a high-dimensional input and compresses it to a latent vector, and $Dec$ uses the latent vector to reconstruct the original input. $Enc$ and $Dec$ are trained as a whole and penalized for creating output that deviates from the input. The loss function is defined as follows:
\begin{equation}
    \underset{\theta, \phi}{min} \ L(\theta, \phi) = \frac{1}{N} \sum_{i=1}^{N} ||x_i - Dec_{\theta}(Enc_{\phi}(x_i))||_{2}^{2},
\end{equation}
where $\theta$ and $\phi$ are the parameters of $Dec$ and $Enc$, $x_{i}$ is a high-dimensional input, $Enc_{\phi}(x_i)$ is the latent vector, $Dec_{\theta}(Enc_{\phi}(x_i))$ is the reconstructed input, and $N$ is the total number of samples.

\label{related_work}

\subsection{Tabular data synthesizers}
We focus on deep-learning approaches for tabular data synthesis 
and skip the discussion of classical methods such as Copulas~\cite{patki2016synthetic, li2020sync} and Bayesian Networks~\cite{zhang2017privbayes}. 
Table~\ref{tab: model_overview} summarizes the recently developed deep learning methods for tabular data synthesis in terms of models, network architecture, and datasets. MedGAN~\cite{choi2017generating} is designed for aggregated electronic health records (EHRs), which only have count and binary features. Since EHRs are high-dimensional and sparse~\cite{baowaly2019synthesizing}, medGAN uses a pre-trained autoencoder to learn compact representations of the input data and thereby simplifies the GAN's task. MedGAN is improved by~\cite{baowaly2019synthesizing}, where the standard GAN loss is replaced by Wasserstein loss with gradient penalty, and the new model is named medWGAN. However, {different from AE-GAN,} medGAN and medWGAN are limited in generalizing to real-world scenarios because they only consider count and binary features. 

A few recent tabular data synthesizers are suitable for general data types, including table-GAN~\cite{tablegan}, CTGAN~\cite{ctgan&tvae}, TVAE~\cite{ctgan&tvae}, and CTAB-GAN~\cite{zhao2021ctab}. CTGAN, TVAE, and CTAB-GAN use Variational Gaussian Mixture (VGM) to encode numerical features and one-hot encoding for categorical features. Moreover, CTAB-GAN defines the \textit{mixed} datatype and proposes a new encoding method. In addition, CTGAN, TVAE, and CTAB-GAN adopt the training-by-sampling technique to handle highly-imbalanced distributions. Despite their effectiveness in tabular data synthesis, these models overlook and do not abide by the key property of column permutation invariance.

\begin{table}[t]
\centering
\caption{Deep learning methods for tabular data synthesis}
\label{tab: model_overview}
\resizebox{\columnwidth}{!}{%
\begin{tabular}{@{}cccc@{}}
\toprule
Method    & Model design                   & Network & Data            \\ \midrule
medGAN~\cite{choi2017generating}    & AE + GAN                       & FCN          & Medical records \\
table-GAN~\cite{tablegan} & DCGAN + Classifier             & CNN          & General         \\
medWGAN~\cite{baowaly2019synthesizing}   & AE + WGAN-GP                   & FCN          & Medical records \\
CTGAN~\cite{ctgan&tvae}     & Conditional WGAN-GP            & FCN          & General         \\
TVAE~\cite{ctgan&tvae}      & Conditional VAE                & FCN          & General         \\
CTAB-GAN~\cite{zhao2021ctab}  & Conditional DCGAN + Classifier & CNN, FCN     & General         \\ \bottomrule
\end{tabular}
}
\end{table}

\subsection{Column permutation invariance} \label{define_column_permutation_invariance}
In computer vision similar concepts 
have been brought up and investigated, including \textit{permutation invariance}~\cite{lee2019set, cohen2020regularizing}, \textit{translation invariance}~\cite{kayhan2020translation, kauderer2017quantifying, furukawa2017deep}, and \textit{translation equivalence}~\cite{weiler2018learning}.
Permutation invariance means that the output of a neural network stays the same despite permutations of its input. For example, the classification of an image should not change after adjusting the object location in the image. 

Motivated by that, we define column permutation invariance in tabular data synthesis as follows. The performance of a tabular data synthesizer should not be affected by permutations on the input column order. To the best of our knowledge, column permutation invariance has not been researched by the prior art on tabular data synthesis. 

\section{Empirical analysis}
In this section, we analyze the column permutation invariance property of the state-of-the-art tabular data synthesizers, with a particular focus on its root causes. 

\subsection{Pitfall of CNNs for tabular data synthesis} \label{cnn_not_a_natural_fit}

\begin{table}[t]
\centering
\caption{TableGAN experiment results: Wasserstein-1 distance between real and synthetic data}
\begin{tabular}{@{}ccccc@{}}
\toprule
\multirow{2}{*}{Dataset} & \multicolumn{3}{c}{Column order}                  & \multirow{2}{*}{\begin{tabular}[c]{@{}c@{}}Max diff. \\ (\%)\end{tabular}} \\ \cmidrule(lr){2-4}
                         & Original order & Order by type  & Order by corr.  &                                                                            \\ \midrule
Loan                              & 2.062                            & \textbf{2.047}                           & 2.066                            & \textbf{0.93\%} \\
Adult                             & 12.153                           & 12.563                          & \textbf{11.512}                           & 9.13\% \\
Credit                            & 0.420                            & 0.410                           & \textbf{0.403}                            & 4.22\% \\
Covtype                           & \textbf{1.282}                            & 1.284                           & 1.345                            & 4.91\% \\
Intrusion                         & 6.486                            & 5.896                           & \textbf{5.645}                            & 14.90\% \\ \midrule
Avg.                              & 4.481                            & 4.440                           & \textbf{4.194}                            & 6.82\% \\ \bottomrule
\end{tabular}
\label{tab:tablegan_sensitivity}
\end{table}

Initially designed for images, CNNs use a set of convolution kernels to slide over the input feature space, abstract high-dimensional features, and then aggregate them into knowledge about the input. Due to the limited kernel size, CNNs only learn local relations between neighboring features and fall short to capture global dependencies.

The focus of CNNs on local relations hinders high quality tabular data synthesis. In contrast to image data, tabular data do not have necessarily strong local relations. Highly-correlated features can be very far apart, and their dependencies can be complex and irregular~\cite{zhu2021converting}. These characteristics make modeling tabular data extra challenging for CNNs~\cite{katzir2021netdnf, pmlr-v97-rahaman19a}, despite their remarkable performance in many machine learning tasks~\cite{sultana2018advancements}.


Since CNNs capture mainly local relations, CNN-based tabular data synthesizers are sensitive to column permutations. We use table-GAN to verify this assumption. We test it with five datasets arranged using three column orders, namely the original order, order by type, and order by correlation. Order by type means putting all continuous columns on the left of the table, and all categorical columns on the right. Order by correlation means placing highly-correlated columns on the left and weakly-correlated columns on the right. For each order, we train the model separately and calculate the Wasserstein-1 distance (WD) between real and synthetic data. 
Every experiment is repeated 5 times. Table~\ref{tab:tablegan_sensitivity} summarizes our results. The best, i.e. lowest, distance values are highlighted in bold. The last column shows the maximum WD change in percent across all three column permutations.
The results show that table-GAN is most sensitive on the Intrusion dataset with a maximum difference in WD of 14.90\%.


\subsection{Sparsity v.s. sensitivity}

Efficient representation of categorical features is one of the main challenges in tabular data synthesis. In the state-of-the-art, table-GAN uses label encoding to transform categorical features into numerical ones and normalizes them with min-max normalization. This method often leads to sub-optimal performance due to the artificial order in categorical features~\cite{hancock2020survey}. In contrast, CTGAN, CTAB-GAN, and TVAE use one-hot encoding to represent categorical features. Despite its simplicity and effectiveness, one-hot encoding introduces many zeros to the input data and thus increases sparsity. 

Representing numerical features in tabular data is relatively straightforward. The most common method is mapping them into [-1, 1] with min-max normalization. However, the authors of CTGAN, TVAE, and CTAB-GAN adopt \textit{mode-specific normalization}, which uses Variational Gaussian Mixture to represent multi-modal numerical features. Although this method improves the quality of the synthetic data, we find it leads to sparse input because one-hot encoding is used to represent multiple modes.

Our experiments show that sparse tabular data causes sensitivity to column permutations. We compare CTAB-GAN, a model using one-hot encoding and mode-specific normalization, with table-GAN, a model using label encoding and min-max normalization. Note that label encoding and min-max normalization do not change the dimensionality of the input data, whereas one-hot encoding and mode-specific normalization increase sparsity. 
Table~\ref{tab:ctabgan_sensitivity} shows our experiment results of CTAB-GAN. Compared to table-GAN (see Table~\ref{tab:tablegan_sensitivity}), CTAB-GAN {synthesizes more realistic data (shown by a lower WD which is about $1/4$ of table-GAN's WD), but with higher} 
sensitivity to column permutations on all datasets.
The average maximum change in WD across the five datasets is 38.67\%, whereas in table-GAN it is 6.82\%.

\begin{table}[t]
\centering
\caption{CTAB-GAN experiment results: Wasserstein-1 distance between real and synthetic data}
\begin{tabular}{@{}ccccc@{}}
\toprule
\multirow{2}{*}{\textbf{Dataset}} & \multicolumn{3}{c}{Column order}                 & \multirow{2}{*}{\begin{tabular}[c]{@{}c@{}}Max diff. \\ (\%)\end{tabular}} \\ \cmidrule(lr){2-4}
                                  & Original order & Order by type  & Order by corr. &                                                                            \\ \midrule
Loan                              & 0.356                            & 0.283                           & \textbf{0.216}                            & 64.81\% \\
Adult                             & 1.517                            & \textbf{0.934}                           & 1.203                            & 62.42\% \\
Credit                            & \textbf{0.115}                            & 0.144                           & 0.137                            & 25.22\% \\
Covtype                           & 0.539                            & \textbf{0.514}                           & 0.583                            & \textbf{13.42\%} \\
Intrusion                         & \textbf{2.668}                            & 3.401                           & 2.831                            & 27.47\% \\ \midrule
Avg.                              & 1.039                            & 1.055                           & \textbf{0.994}                            & 38.67\% \\ \bottomrule
\end{tabular}
\label{tab:ctabgan_sensitivity}
\end{table}

\begin{figure}[t]
  \centering
    \includegraphics[width=.5\linewidth]{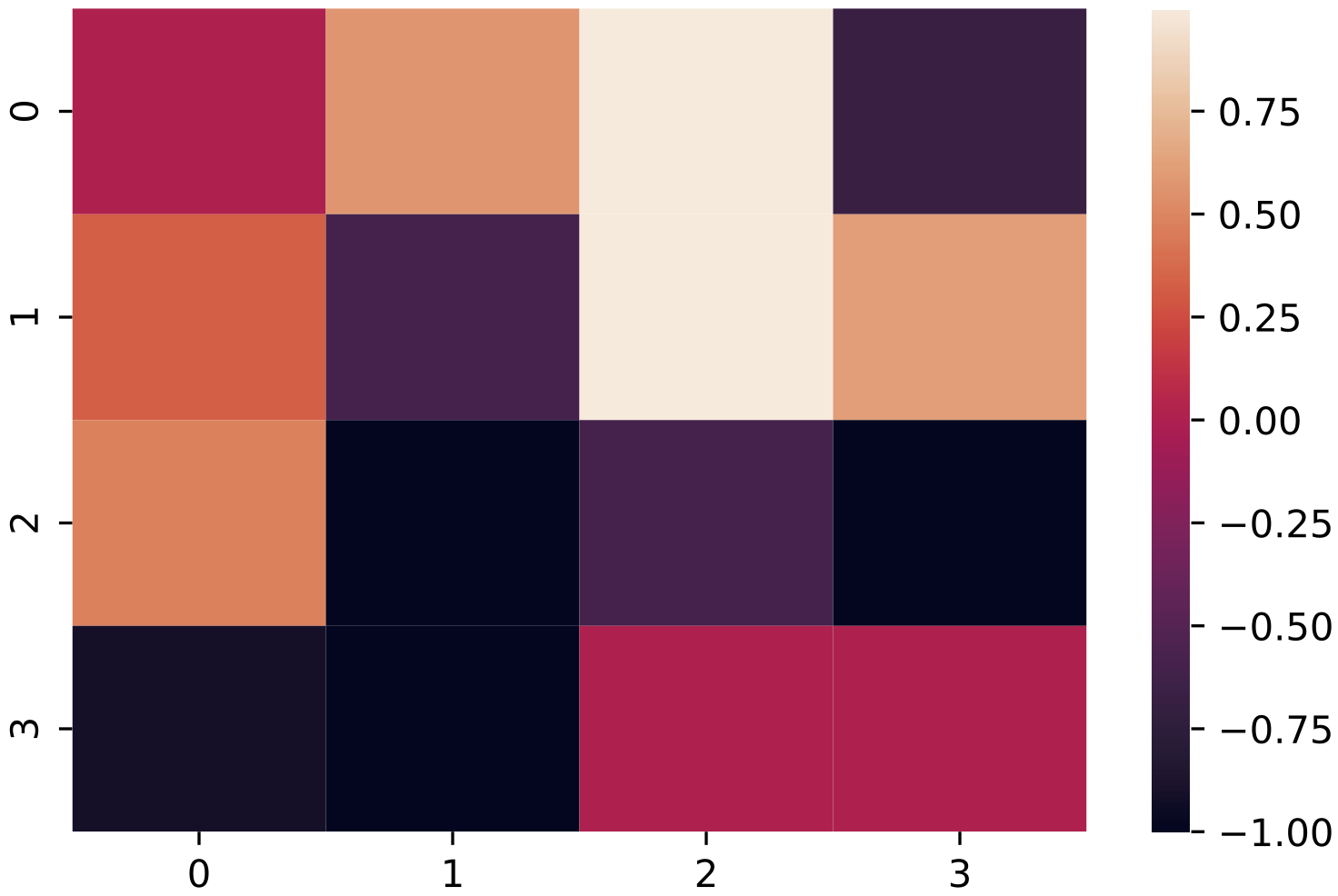}\hfill
    \includegraphics[width=.5\linewidth]{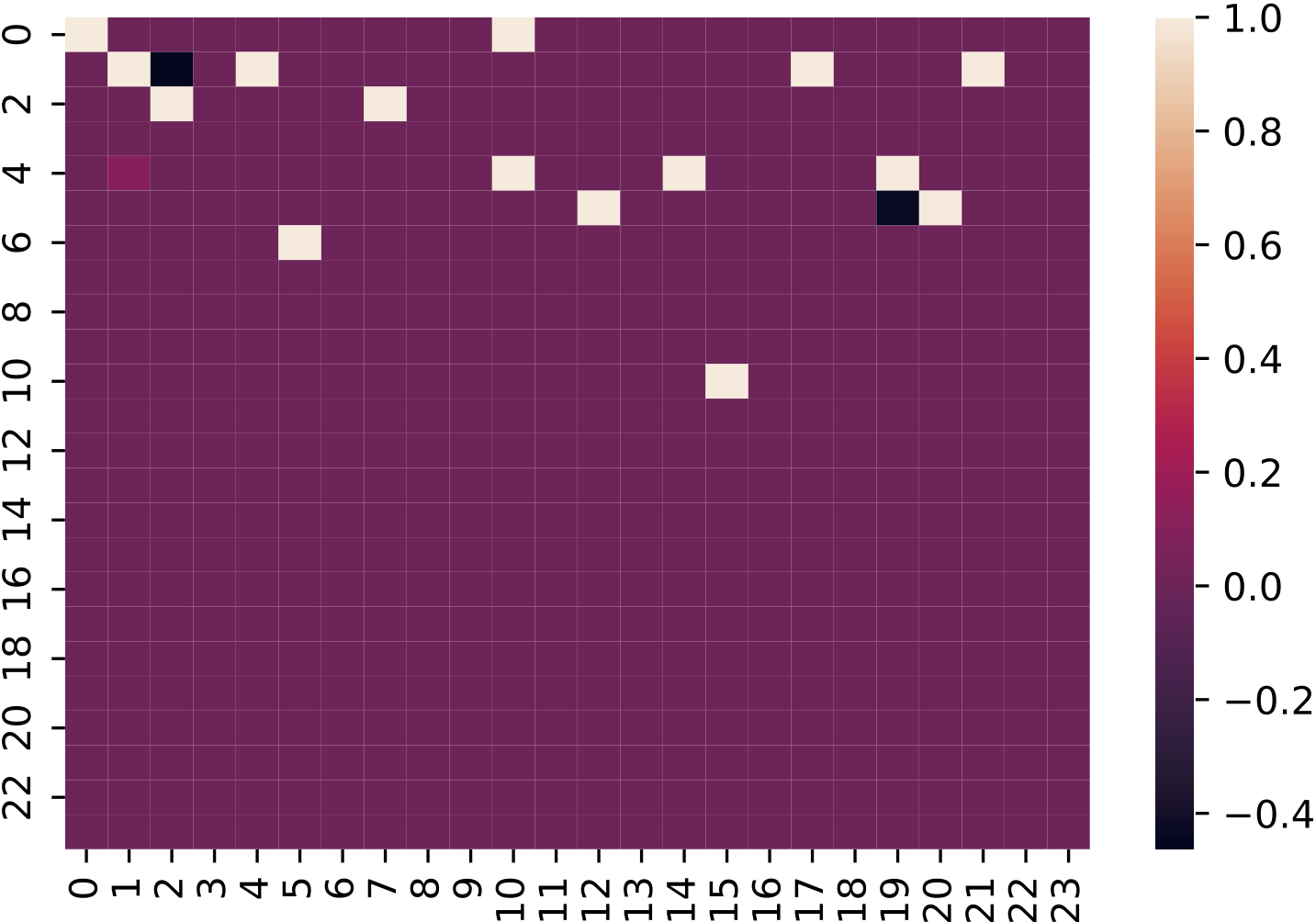}
    \caption{Visualization of the encoded input to table-GAN (left) and to CTAB-GAN (right) using Adult dataset.}
    \label{fig: visualize_input_to_tablegan&ctabgan}
\end{figure}

To further highlight the sparsity of the encoded data we visualize the input of table-GAN and CTAB-GAN in Figure~\ref{fig: visualize_input_to_tablegan&ctabgan}. We note that the sparsity is determined by the sum of all levels of all variables, i.e., the number of modes per continuous variable and the discrete levels per discrete variable.  We reshape each row of a table into a square matrix to make it compatible with CNNs. In table-GAN, one row of the Adult dataset is represented by a $4 \times 4$ matrix, but in CTAB-GAN a $24 \times 24$ matrix is needed due to one-hot encoding and mode-specific normalization which increases the number of zeros, i.e. purple matrix cells in the figure accounting for roughly 97\% area. This increase in sparsity makes CTAB-GAN more sensitive to column permutations than table-GAN since the average distance between related columns (pixels) is increased.

\subsection{FCNs column permutation invariance}
Theoretically, fully-connected networks (FCNs) should be robust to column permutations because all features are connected. However, we find FCNs are not fully permutation invariant. Tests with CTGAN and TVAE, two FCN-based tabular data synthesizers, on five datasets show an average WD of $1.87$ with an average maximum change of 18.62\% for CTGAN and $1.80$ with 14.89\% for TVAE (details skipped due to space constraints).
{Overall, the state-of-the-art tabular data synthesizers either provide the high quality synthetic data or are resilient to column order permutations, struggling to make a good trade-off.}

\section{AE-GAN}

\label{ae-gan}
\begin{figure}[t]
    \centering
    \includegraphics[width=1\linewidth]{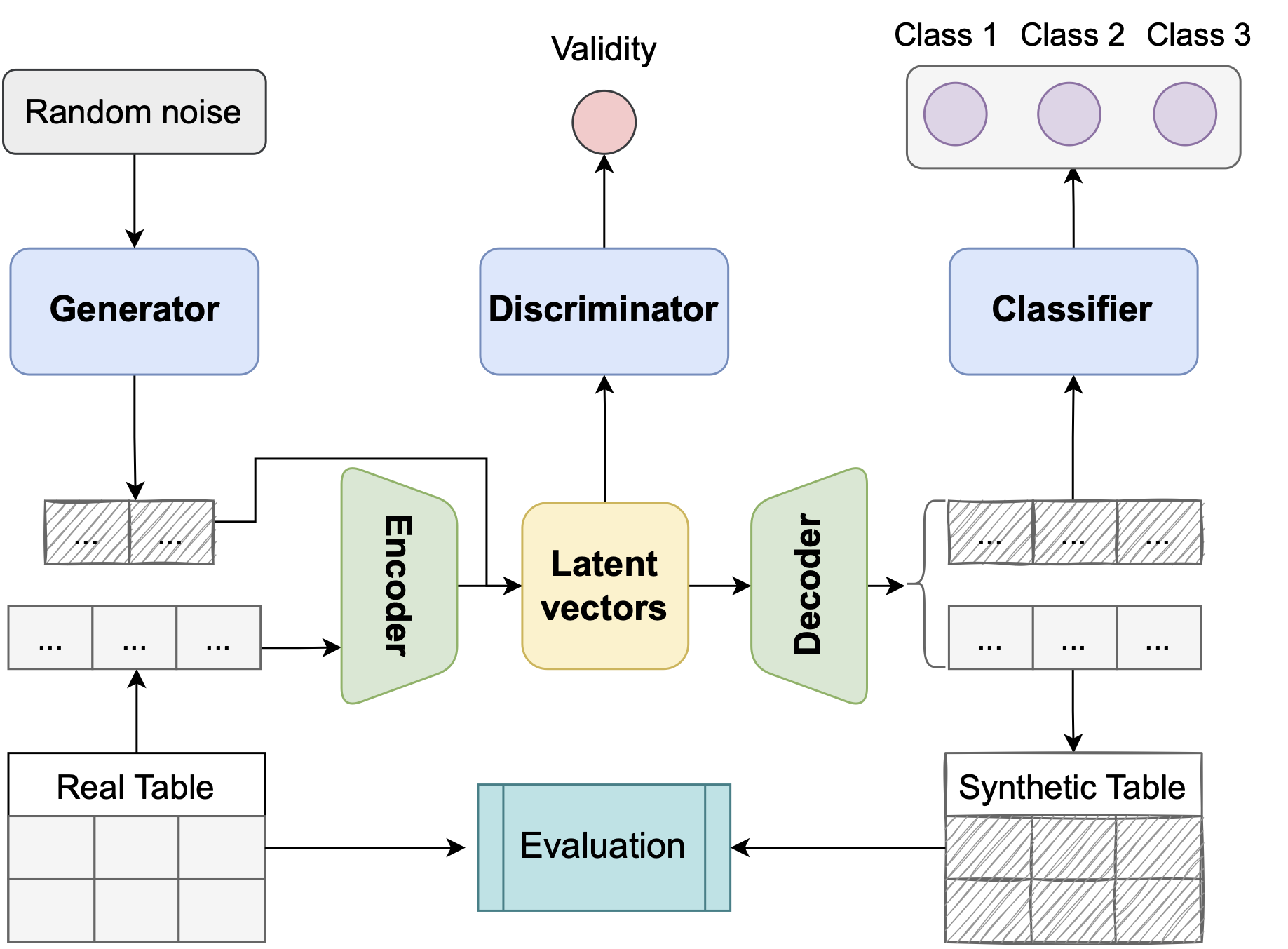}
    \caption{The overall architecture and data flow of AE-GAN.}
    \label{fig:ae-gan}
\end{figure}

We propose AE-GAN, a GAN-based tabular data synthesizer, which aims to improve the resilience to the input column permutations by using 
latent representations of tabular data via an autoencoder. Figure~\ref{fig:ae-gan} shows the overall architecture and data flow of AE-GAN. It has five components: Encoder ($Enc$), Decoder ($Dec$), Generator ($G$), Discriminator ($D$), and Classifier ($C$). The main objective of the encoder and the decoder is to find a {more compact} latent representation of the input data, which follow the data encoding scheme proposed below. Once the autoencoder is trained, the encoded latent vector and the random noise vector are used as input to train the GAN. The GAN aims to generate a synthetic latent vector having high similarity to the original one. During training the classifier provides additional feedback to ensure the semantic integrity of synthetic data. We explain the design choice of each component in the following. 

\subsection{Data representation}
Following~\cite{ctgan&tvae}, we use mode-specific normalization for numerical features and one-hot encoding for categorical features. Mode-specific normalization preserves the multi-modal distribution of numerical features and improves the performance of tabular data synthesizers~\cite{ctgan&tvae, zhao2021ctab}. One-hot encoding is a simple yet effective way to convert categorical values to numerical ones without losing too much information.
Certainly, one-hot encoding and mode-specific normalization causes sparse input, but the autoencoder in AE-GAN maps the sparse encoded input into compact latent vectors solving this issue. 

\subsection{Encoder and decoder}
Since we identify sparsity as one of the main reasons for tabular data synthesizers' sensitivity to column permutations, one natural solution is to use an autoencoder to extract the features of tabular data and compress them into compact latent vectors. Such an advantage can apply to all kinds of tabular data synthesizers. 

We use two three-layer fully connected networks as $Enc$ and $Dec$. Based on our study of mainstream open-source implementations of autoencoders~\cite{liao_2020, zhao_2020, zhou_2020}, fully-connected networks with 2-4 layers are common choices for autoencoders.

Another important design choice is the length of the latent vector, i.e., the output size of $Enc$ and the input size of $Dec$. This determines the autoencoder's capacity to represent high-dimensional data. We choose this parameter based on the size of the input dataset. For datasets with a large number of columns, we increase the length of the latent vector to ensure that  complex relations between columns can be well represented, therefore helping the generator to synthesize realistic data. 

\subsection{Generator and discriminator}
\begin{figure}[t]
    \centering
    \includegraphics[width=1\linewidth]{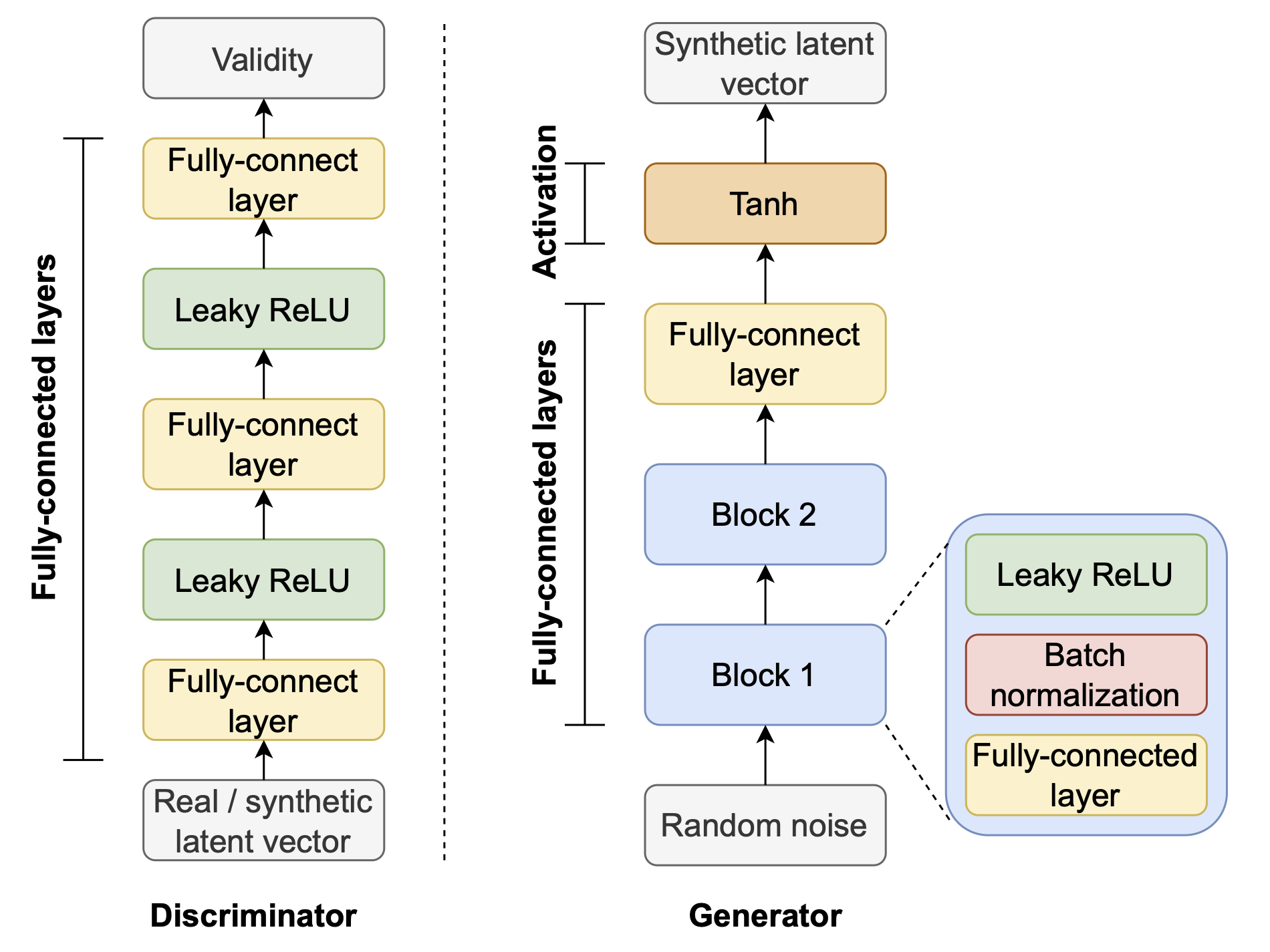}
    \caption{Architectures of the discriminator and the generator in AE-GAN.}
    \label{fig:ae_gan_architecture_g&d}
\end{figure}

The core of AE-GAN is a GAN, which has two competing networks, namely the generator, $G$, and the discriminator, $D$.
Figure~\ref{fig:ae_gan_architecture_g&d} shows their architectures.  
Both $G$ and $D$ have three fully-connected layers which are more resilient to {permutations of elements in the (more compact) latent vector.} 
In the discriminator, the fully connected layers are followed by leaky ReLU activation and the final output is the validity of the input. In the generator, the fully-connected layers are followed by batch normalization and leaky ReLU activation, and the final output is the synthetic latent vector. The design is based on the architecture of WGAN-GP~\cite{wgan_gp_github_2019}.

\subsection{Auxiliary classifier}
To enhance the training of $G$ and thus the quality of the generated tabular data, we introduce an auxiliary classifier $C$ to the GAN. This design is inspired by~\cite{tablegan}, where an auxiliary classifier is added to maintain the semantic consistency of synthetic data. Note that the input to $C$ is the reconstructed data from $Dec$, rather than the latent data from $Enc$ and $G$ because we want $C$ to learn the semantic relations between columns directly. Specifically, given a categorical target column with several categories, $C$ learns to classify which category a sample belongs to according to the columns other than the target column. This is how $C$ differs from the $D$: $D$ determines the ``realness'' of a sample based on all columns in the latent space, whereas $C$ learns the relationship between the target column and all other columns in the reconstructed space. By combining $C$ with the GAN, we simultaneously leverage the flexibility of unsupervised training with the control provided by supervised training, thereby improving the quality of the synthetic data.

\subsection{Loss functions}
The training of AE-GAN requires four loss functions: autoencoder loss $\mathbb{L}_{AE}$, generator loss $\mathbb{L}_{G}$, discriminator loss $\mathbb{L}_{D}$, and classifier loss $\mathbb{L}_{C}$. 

\subsubsection{Autoencoder loss}
Autoencoder loss is the reconstruction loss, i.e., the element-wise mean squared error between its input and reconstructed output. It is defined as follows:
\begin{equation}
    \mathbb{L}_{AE} = \mathbb{E}||x-\tilde{x}||_2^2,
\end{equation}
where $x$ and $\tilde{x}$ are the input and the reconstructed output.

\subsubsection{Generator loss}
The generator receives feedback from both the discriminator and the classifier. Therefore, its loss function is the sum of: discriminator feedback $\mathbb{L}_{G}^{D}$ and classifier feedback $\mathbb{L}_{G}^{C}$. 
\begin{equation}
    \mathbb{L}_{G} = \mathbb{L}_{G}^{D} + \mathbb{L}_{G}^{C}
\end{equation}

Discriminator feedback is the validity of synthetic samples:
\begin{equation}
    \mathbb{L}_{G}^{D} = - \mathbb{E}[D(G(z))],
\end{equation}
where $G(z)$ is the generator output and $D(G(z))$ is the discriminator output.

Classifier feedback is the cross entropy between the predicted value and the actual value of the target column:
\begin{equation}
    \mathbb{L}_{G}^{C} = H(m,m'),
\end{equation}
where $m$ and $m'$ are the actual and predicted values of the target column, and $H(\cdot)$ is the cross entropy operator.

\subsubsection{Discriminator loss}
The discriminator loss measures how well it differentiates the real samples and the synthetic samples. We use Wasserstein loss with gradient penalty to improve the training stability and alleviate the vanishing gradient problem of GANs~\cite{gulrajani2017improved}. It is calculated by:
\begin{equation}
    \mathbb{L}_{D} = -\mathbb{E}[D(x) - D(G(z)) - \lambda \cdot (||\nabla D(\hat{x})||_2-1)^2],
\end{equation}
where $D(x), D(G(z))$ and $D(\hat{x})$ are the discriminator output on real samples, synthetic samples, and the interpolates between real and synthetic samples. $\lambda$ is the gradient penalty coefficient, $\nabla D(\hat{x})$ is the gradient of $D(\hat{x})$ on $\hat{x}$.

\subsubsection{Classifier loss}
The classifier loss also has two parts: loss on real samples $\mathbb{L}_{C}^{R}$ and loss on synthetic samples $\mathbb{L}_{C}^{S}$. 
\begin{equation}
    \mathbb{L}_{C} = \mathbb{L}_{C}^{R} + \mathbb{L}_{C}^{S}.
\end{equation}
The calculation of $\mathbb{L}_{C}^{R}$ and $\mathbb{L}_{C}^{S}$ are similar to $\mathbb{L}_{G}^{C}$.

\subsection{Training algorithm}
{We test two training strategies: disjoint training and joint training. For disjoint training
we first train the AE until convergence and then train the GAN with the classifier while utilizing the compression power of the AE.
For joint training, inspired by TimeGAN~\cite{yoon2019time} and the hypothesis of possible training synergies, 
we first pre-train the autoencoder for a certain number of epochs and then co-train it with the GAN and the classifier.
Ablation tests, details in Section~\ref{ssec:ablation}, show that disjoint training achieves lower training losses. Hence we use disjoint training for all results if not specified.}


\section{Feature Sorting Algorithm}
\label{feature_sorting_algorithm}
Although features in tabular data are often heterogeneous, some are highly correlated. 
However, as discussed in Section~\ref{cnn_not_a_natural_fit}, CNN-based tabular data synthesizers often fail to capture these correlations due to the distance between correlated columns.

A simple solution for this problem is to put highly-correlated features together, such that a convolution kernel can capture them simultaneously. 

The location of features also matters. 
In a convolution process, features on the border of the input matrix are convoluted fewer times than features within the border, leading to potential loss of information. 
This is called \textit{boundary effects} in image processing~\cite{castleman1996digital, strang1996wavelets} which can lead to statistical biases in finite-sampled data~\cite{kayhan2020translation, griffith1983boundary, griffith1983evaluation}.
To better capture the correlation between features, one must also carefully choose the location of high-correlated features. 

\begin{figure}[t]
    \centering
    \includegraphics[width=1\linewidth]{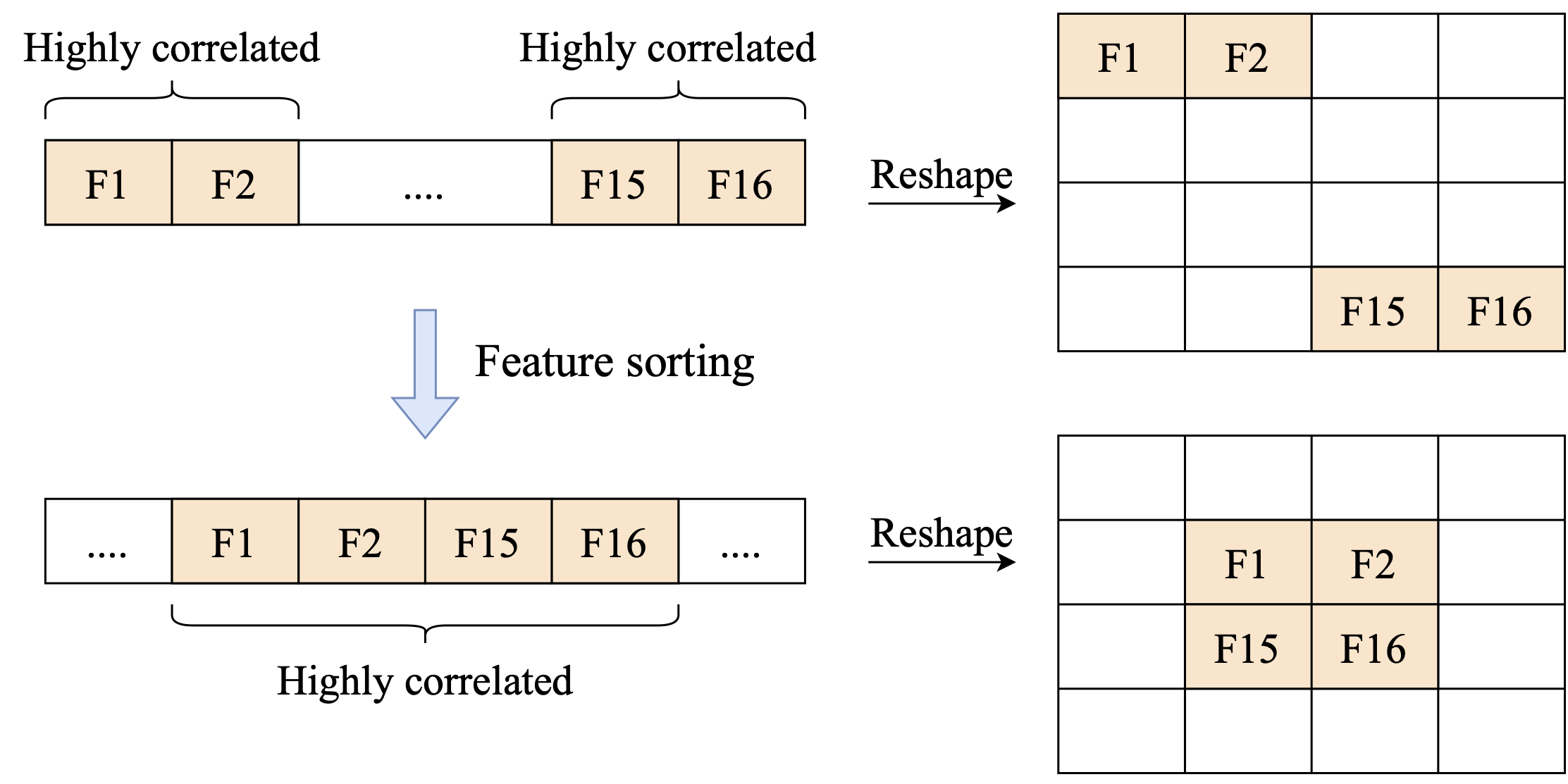}
    \caption{\textbf{Top: }The highly correlated features, $F1, F2, F15$ and $F16$ are on the border of the input matrix, suffering from the boundary effects of CNNs. \textbf{Bottom: }After feature sorting, $F1, F2, F15$ and $F16$ are at the center of the input matrix.}
    \label{fig:put_high_corr_cols_in_mid}
\end{figure}

To address both issues our solution is to group highly-correlated features together in the middle of the table and mitigate the boundary effect. 
Figure~\ref{fig:put_high_corr_cols_in_mid} illustrates the base of our idea. Since each row must be reshaped into a square matrix to be fed into a CNN, we have to carefully sort the features such that they end up in the middle after reshaping. 

Although this idea seems straightforward, one complication arises due to the encoding of features. Variational Gaussian Mixture~\cite{ctgan&tvae} and One-hot encoding require multiple columns to represent one feature. Consequently, each feature may occupy a different number of columns. To put highly correlated features in the middle after encoding, we must consider the length of each encoded feature.

We developed a feature sorting algorithm that groups  highly-correlated features and puts them in the middle, see Algorithm~\ref{alg:feature_sort}. 
We first pick the most correlated feature pairs and then add other features to their left or right side. We have two counters, $c_{left}$ and $c_{right}$, for columns taken by features added to the left and the right side. Before adding a feature, we compare $c_{left}$ and $c_{right}$, and then add it to the side with fewer columns. Thereby we ensure that the highly correlated features stay in the middle even after encoding.

\begin{algorithm}[t]
\caption{Feature Sorting Algorithm} 
\label{alg:feature_sort}
\begin{algorithmic}[1]
 \renewcommand{\algorithmicrequire}{\textbf{Input:}}
 \renewcommand{\algorithmicensure}{\textbf{Output:}}
 \Require Original Table $T_o = \{F_0, F_1, ..., F_n\}$
 \Ensure  Sorted Table $T_{sorted}$
   \State $T_{sorted} \gets \{\}$
   \State $c_{left}, c_{right} \gets 0$ \Comment{No. columns added to the left / right of $T_{sorted}$}
   \State $corr \gets [~]$ \Comment{Pair-wise correlation / association}
   \For {all possible pairs of features in $T_o$}
   \State Calculate the absolute value of their correlation / association and save it in $corr$
   \EndFor
   \While {$length(T_{sorted}) \neq length(T_{o})$ }
   \State Find the largest value $v$ in $corr$
   \State Find the corresponding pair of features $\{F_x, F_y\}$
   \State $F_{new} \gets \{F_x, F_y\} - \{F_x, F_y\} \cap T_{sorted}$ \Comment{Feature(s) not yet in $T_{sorted}$}
   \State $c \gets$ No. columns occupied by $F_{new}$ after encoding
    \If {$T_{sorted}$ is empty}
    \State $T_{sorted} \gets \{F_x, F_y\}$ \Comment{Add the first pair}
    \Else
    \If {$c_{right} < c_{left}$}
    \State $T_{sorted} \gets T_{sorted} + F_{new}$ \Comment{Add the new feature(s) to the right}
    \State $c_{right} \gets c_{right} + c$
    \Else
    \State $T_{sorted} \gets  F_{new} + T_{sorted}$ \Comment{Add the new feature(s) to the left}
    \State $c_{left} \gets c_{left} + c$
    \EndIf
    \EndIf
    \State Remove $v$ from $corr$
   \EndWhile
 \State \Return $T_{sorted}$ 
 \end{algorithmic} 
\end{algorithm}

\section{Experimental evaluation}
\subsection{Experimental Setup}

\subsubsection{Datasets}

We use five tabular datasets that are common in the machine learning community. Table~\ref{tab: dataset_overview} summaries their main statistics. The Loan dataset\footnote{https://www.kaggle.com/code/pritech/bank-personal-loan-modelling/data} contains the demographic information about bank customers and their response to a personal loan campaign. The Adult dataset\footnote{https://archive.ics.uci.edu/ml/datasets/Adult} has many census data and is used to predict whether the income of an adult exceeds \$50k/year. The Credit dataset\footnote{https://www.kaggle.com/datasets/mlg-ulb/creditcardfraud}
consists of anonymized credit card transactions labeled as fraudulent or genuine. 
The Intrusion dataset\footnote{https://archive.ics.uci.edu/ml/datasets/Unmanned+Aerial+Vehicle+\%28U\\AV\%29+Intrusion+Detection}
has encrypted WiFi traffic records and classifies whether a record is from an unmanned aerial vehicle. The Covtype dataset\footnote{https://archive.ics.uci.edu/ml/datasets/Covertype} contains the cover type of forests and the related geographical information. Every dataset has a target column for classification tasks. Due to the limitation of computational resources, we randomly select 50k samples from the Credit, Intrusion, and Covtype datasets.


\begin{table}[t]
\centering
\caption{Statistics of datasets}
\begin{tabular}{@{}ccccc@{}}
\toprule
Datasets  & \begin{tabular}[c]{@{}c@{}}\# Continuous\\ columns\end{tabular} & \begin{tabular}[c]{@{}c@{}}\# Categorical \\ columns\end{tabular} & \begin{tabular}[c]{@{}c@{}}\# Columns\\  after encoding\end{tabular} & \# Samples \\ \midrule
Loan      & 5                                                               & 8                                                                 & 55                                                                  & 5k         \\
Adult     & 5                                                               & 9                                                                 & 151                                                                 & 48k        \\
Credit    & 30                                                              & 1                                                                 & 301                                                                 & 50k        \\
Covertype & 10                                                              & 45                                                                & 205                                                                 & 50k        \\
Intrusion & 22                                                              & 20                                                                & 322                                                                 & 50k        \\ \bottomrule
\end{tabular}
\label{tab: dataset_overview}
\end{table}

\subsubsection{Baselines}

Four state-of-the-art tabular data synthesizers are selected as the baseline models, namely table-GAN~\cite{tablegan}, CTGAN~\cite{ctgan&tvae}, TVAE~\cite{ctgan&tvae}, and CTAB-GAN~\cite{zhao2021ctab}. We use the same hyperparameters as the original papers, and every experiment is repeated three times to obtain reliable results. 

\subsubsection{Computational Environment}

We implemented our proposed solutions using Pytorch on a server equipped with an Intel(R) Core(TM) i9-10900KF CPU @3.70GHz and a GeForce RTX 2080 Ti GPU.

\subsection{Evaluation Metrics}
Our evaluation of tabular data synthesizers focuses on the statistical difference and machine learning utility difference between real and synthetic data. Two metrics quantify the statistical difference.

{\bf Wasserstein-1 Distance (WD)} measures the difference between two continuous/discrete 1-dimensional distributions. We use this metric to compare the per features difference between real and synthetic data. 

{\bf Difference in Correlation Matrix (Dif. Corr.)} 
measures how well the cross-column correlations\footnote{Note that we use ``correlation'' as a general term. For two numerical features, it refers to the Pearson correlation coefficient; for two categorical features, it is their Cramer's V; and for a categorical features and a numerical features, it means their correlation ratio.} are captured by a tabular data synthesizer.
We calculate the difference between the correlation matrices of the real and synthetic table as follows:
\begin{equation}
    Dif. \ Corr. = \sqrt{\sum_{i,j}(Corr^{R}_{i,j} - Corr^{F}_{i,j})^2},
\end{equation}
where $Corr^{R}_{i,j}$ and $Corr^{F}_{i,j}$ are the correlation coefficients between features $i$ and $j$ in the real and synthetic correlation matrices.

We measure machine learning utility as the performance differences of machine learning models trained on the real and synthetic data. Specifically, we first train four machine learning models with real and synthetic data separately. 
Then we obtain their average prediction accuracy and compute the difference.
The difference is small if the synthetic data has high machine learning utility. 

\subsection{AE-GAN}
AE-GAN is evaluated on four aspects: column permutation invariance, statistical similarity and  machine learning utility of the synthetic data, and training time. 
We aim to verify if AE-GAN is robust to column permutations and achieves good synthesis quality compared with the state-of-the-art. Additionally, we evaluate the scalability of AE-GAN by analyzing its training time. Table~\ref{tab:all_result_in_one} summarises the results averaged on five datasets.

\begin{table}[t]
\centering
\caption{AE-GAN evaluation results against the state-of-the-art. For all metrics, a lower value is better.}
\resizebox{0.5\textwidth}{!}{%
\begin{tabular}{@{}cccccc@{}}
\toprule
\multirow{2}{*}{Model} & \multirow{2}{*}{\begin{tabular}[c]{@{}c@{}}Sensitivity to \\ permutations\end{tabular}} & \multicolumn{2}{c}{\begin{tabular}[c]{@{}c@{}}Stat. diff. \end{tabular}} & \multirow{2}{*}{\begin{tabular}[c]{@{}c@{}}ML utility\\ diff. \end{tabular}} & \multirow{2}{*}{\begin{tabular}[c]{@{}c@{}}Training time \\ (mins)\end{tabular}} \\ \cmidrule(lr){3-4}
                      &                                                                                                & WD                                                        & Dif. Corr.                                                &                                                                                                                   &                                                                                  \\ \midrule
table-GAN              & \textbf{6.82\%}                                                                                & 4.481                                                     & 3.651                                                     & 21.14\%                                                                                                           & \textbf{1.31}                                                                    \\
CTAB-GAN               & 38.67\%                                                                                        & \textbf{1.039}                                            & \textbf{1.905}                                            & \textbf{9.11\%}                                                                                                   & 63.70                                                                            \\
CTGAN                  & 18.62\%                                                                                        & 1.857                                                     & 3.079                                                     & 14.99\%                                                                                                           & 10.47                                                                            \\
TVAE                   & 17.29\%                                                                                        & 1.723                                                     & 2.848                                                     & 12.84\%                                                                                                           & 7.00                                                                             \\ \midrule
AE-GAN                 & 11.71\%                                                                                        & 2.699                                                     & 2.331                                                     & 9.98\%                                                                                                            & 9.89                                                                             \\ \bottomrule
\end{tabular}
}
\label{tab:all_result_in_one}
\end{table}

\textbf{Column Permutation Invariance}. Similar to our empirical analysis, we arange training data in three different orders, i.e., original order, order by type, and order by correlation, and test the performance of AE-GAN. The second column of Table \ref{tab:all_result_in_one} shows the sensitivity to column permutations of the baseline models and AE-GAN, averaged on five datasets. We find AE-GAN ranks second in permutation invariance among the five models. Table-GAN is the most permutation-invariant model because it does not have the sparsity issue caused by one-hot encoding and mode-specific normalization. In contrast, CTAB-GAN, TVAE, CTGAN and AE-GAN adopt one-hot encoding for categorical features and mode-specific normalization for numerical features and thus have sparse input, leading to higher sensitivity. However, since AE-GAN has an autoencoder to compress the input, it is more robust to column permutations than CTAB-GAN, TVAE, and CTGAN.

\textbf{Synthesis Quality Comparison}. We evaluate the quality of the synthesized data with two metrics: statistical difference and ML utility difference between real and synthetic data. Table~\ref{tab:all_result_in_one} shows that CTAB-GAN is the best model in synthesis quality because its synthetic data have the lowest statistical difference and ML utility difference compared with real data. Table-GAN is the worst among all models. In the rest three models, AE-GAN is better than CTGAN and TVAE on Dif. Corr, but worse on WD. However, the ML utility of AE-GAN is better than CTGAN and TVAE. The results show that the autoencoder of AE-GAN helps preserve the correlation between different features, but for each feature, the statistical difference between real and synthetic data may increase due to the information loss caused by data compression. {Future work may look into optimizing the trade-off between the performance boost for GAN and the compression loss caused by compact representations.}


\textbf{Training Time Analysis}. A model with a short training time can scale up to large datasets. It also requires fewer hardware resources than slow models given the same input. We compare the scalability of AE-GAN with the baseline models by analyzing their total training time. Table~\ref{tab:all_result_in_one} shows that AE-GAN is faster than CTGAN and CTAB-GAN, but slower than table-GAN and TVAE. Table-GAN has the shortest training time because of its simple data representation, which leads to small input size. TVAE is faster than AE-GAN because AE-GAN has an auxilliary classifier. Nonetheless, AE-GAN is significantly faster than CTAB-GAN, speeding up as much as 6 times.

\textbf{Summary}. AE-GAN achieves the best tradeoff between permutation invariance, synthesis quality, and training time compared with state-of-the-art tabular data synthesizers. It is more permutation-invariant than CTAB-GAN, CTGAN, and TVAE, and it has better synthesis quality than table-GAN, CTGAN, and TVAE in terms of ML utility. Although table-GAN is less sensitive to column permutations and takes less time to train than AE-GAN, its synthesis quality is much worse. In a similar vein, although CTAB-GAN is better than AE-GAN in synthesis quality, it is much slower and more sensitive to column permutations. Compared with CTGAN and TVAE, AE-GAN is more permutation-invariant, has better ML utility, and takes a similar time to train.

\subsection{Ablation Study}
\label{ssec:ablation}
We conduct an ablation study to understand the influence of the design choices we made with AE-GAN. We change the data representations, model architecture, and training algorithm to test their effect. Table~\ref{tab: ablation study} summarizes the results of the ablation study. 

\begin{table*}[htp]
\centering
\caption{Ablation study results on synthesis quality and permutation invariance of AE-GAN}
\label{tab: ablation study}
\resizebox{\textwidth}{!}{%
\begin{tabular}{@{}ccccccccc@{}}
\toprule
\multirow{2}{*}{Dataset} & \multicolumn{5}{c}{WD between real and synthetic data}                      & \multicolumn{3}{c}{Sensitivity to column permutations} \\ \cmidrule(l{4pt}r{4pt}){2-6} \cmidrule(l{4pt}r{4pt}){7-9}
                         & AE-GAN & w/o MSN & w/o one-hot \& MSN & w/o classifier & co-train AE \& GAN & AE-GAN        & w/o MSN      & w/o one-hot \& MSN      \\ \midrule
Loan                     & 1.374  & 2.309   & 3.749              & 1.308          & 1.880              & 7.28\%        & 3.29\%       & 4.45\%                  \\
Adult                    & 6.042  & 6.319   & 15.432             & 6.293          & 6.508              & 21.77\%       & 39.20\%      & 40.65\%                 \\
Credit                   & 0.341  & 1.650   & 1.505              & 0.344          & 0.534              & 9.09\%        & 3.19\%       & 2.70\%                  \\
Covtype                  & 1.408  & 3.099   & 5.954              & 1.428          & 2.437              & 10.16\%       & 5.48\%       & 0.79\%                  \\
Intrusion                & 4.328  & 14.915  & 58.241             & 4.492          & 4.836              & 10.28\%       & 26.84\%      & 2.76\%                  \\ \midrule
Avg.                     & 2.699  & 5.658   & 16.976             & 2.773          & 3.239              & 11.71\%       & 15.60\%      & 10.27\%                 \\ \bottomrule
\end{tabular}}
\end{table*}

\textbf{Without Mode-Specific Normalization (MSN)}. We use mode-specific normalization in AE-GAN to normalize numerical features. Although it preserves the multi-model distribution of numerical features, it increases the sparsity in training data. We replace it with min-max normalization to understand its effect. Table~\ref{tab: ablation study} shows that after removing mode-specific normalization, the WD becomes worse on all datasets, meaning that mode-specific normalization improves the synthesis quality. However, it also makes AE-GAN more sensitive to column permutations. After removing it, the sensitivity to column permutations decreases on the Loan, Credit, and Covtype datasets. In conclusion, mode-specific normalization increases sensitivity to column permutations, but it improves synthesis quality.

\textbf{Without One-hot and Mode-Specific Normalization}.
To further reduce sparsity in the input data, we remove one-hot encoding and mode-specific normalization together. Similar to table-GAN, we pre-process categorical and numerical features using min-max normalization. We found that the WD is worse than only removing mode-specific normalization, which proves that one-hot encoding can enhance synthesis quality. Moreover, the sensitivity to column permutations decreases on all datasets except the Adult dataset after removing one-hot encoding and mode-specific normalization, especially on datasets with a high proportion of categorical features such as the Covtype and Intrusion datasets. The results verify again that reducing sparsity can enhance permutation invariance. 

\textbf{Without Auxiliary Classifier}. We use an auxiliary classifier to improve the synthesis quality of AE-GAN. After removing the auxiliary classifier, the Wasserstein distance between real and synthetic data worsens on all datasets except the Loan dataset. Overall, the average WD on five datasets increases from $2.669$ to $2.773$ after removing auxiliary classifier, showing that the classifier improves synthesis quality.

\textbf{Co-training AE and GAN}. The AE and GAN in AE-GAN are trained separately. To study whether co-training AE and GAN can improve the synthesis quality, we first pre-train the AE for 300 epochs and then train it together with GAN. Surprisingly, the results show that co-training makes the synthesis quality worse. We find that the training loss of AE is already low after pre-training. However, during co-training, the feedback from GAN increases AE's loss and makes it unstable. 

\subsection{Feature sorting algorithm}
We evaluate the proposed feature sorting algorithm on table-GAN and CTAB-GAN, two CNN-based tabular data synthesizers, because this algorithm is designed to alleviate the limitations of CNN as explained in Section~\ref{feature_sorting_algorithm}.

\textbf{Table-GAN}. Table~\ref{tab: tablegan-feature-sorting} shows the effect of the feature sorting algorithm on table-GAN. A negative change in Dif. Corr. or WD means the difference between synthetic and real data becomes smaller. That is, the feature sorting algorithm helps tabular data synthesizers generate more realistic data. The results show that the feature sorting algorithm works best on the Credit dataset, where Dif. Corr. and WD are decreased by 12\% and 4\%. It also improves the results on the Intrusion dataset, where WD is reduced by 16\%, whereas Dif. Corr slightly increases by 3\%. However, it does not influences much the Loan and Covtype datasets, where the Dif. Corr. and WD change less than 5\%. Moreover, the results on the Adult dataset become worse, with Dif. Corr and WD increase by 24\% and 3\%. 

The algorithm performs best on the Credit dataset because of the simple correlations between its features. Using $\displaystyle \pm 0.2$ as the threshold for high correlation, only \textit{Time} and \textit{Amount} are strongly-correlated with other features. All other features have a close-to-0 correlation. Besides, \textit{Time} and \textit{Amount} are only correlated with 3 and 5 features, respectively. With such a small number of correlated features, capturing their relation in the convolution process is easy once we group them together. 

The algorithm also alleviates the CNN boundary effect on the highly-correlated features of the Credit dataset. In the original order, \textit{Time} and \textit{Amount} are the leftmost and rightmost columns in the table, and many of their correlated features are far apart. However, after feature sorting, these features are in the middle of the table therefore reducing the boundary effect.

In contrast to the Credit dataset, the other four datasets have a larger number of correlated features. For example, in the Adult dataset most features are correlated with at least one other feature, and seven features are correlated with more than three features. Due to the limited kernel size, it is challenging for CNNs to capture all the cross-column relations even after putting the highly-correlated features together. Besides, our algorithm is based on pairwise correlation, but putting one pair of highly-correlated features together could possibly separate another pair of highly-correlated features, which explains why sometimes Dif. Corr. and WD become worse after applying the feature sorting algorithm. In this case, domain knowledge is required to effectively group the correlated features and arrange them in a good order. 

\begin{table}[t]
\centering
\caption{Table-GAN before and after the feature sorting algorithm.
}
\label{tab: tablegan-feature-sorting}
\begin{tabular}{@{}ccccccc@{}}
\toprule
\multirow{2}{*}{Dataset} & \multicolumn{2}{c}{Before sorting} & \multicolumn{2}{c}{After sorting} & \multicolumn{2}{c}{Change} \\ 
\cmidrule(l{4pt}r{4pt}){2-3}
\cmidrule(l{4pt}r{4pt}){4-5}
\cmidrule(l{4pt}r{4pt}){6-7}
                                  & Dif. Corr.      & WD      & Dif. Corr.      & WD     & Dif. Corr.  & WD  \\ \midrule
Loan                              & 2.284                    & 2.062            & 2.203                    & 2.087           & -4\%                 & 1\%          \\
Adult                             & 1.563                    & 12.153           & 1.942                    & 12.502          & 24\%                 & 3\%          \\
Credit                            & 3.092                    & 0.420             & 2.728                    & 0.403           & -12\%                & -4\%         \\
Covtype                           & 4.885                    & 1.282            & 4.915                    & 1.348           & 1\%                  & 5\%          \\
Intrusion                         & 6.433                    & 6.486            & 6.597                    & 5.418           & 3\%                  & -16\%        \\ \midrule
Avg.                              & 3.651                    & 4.481            & 3.677                    & 4.352           & 1\%                  & -3\%         \\ \bottomrule
\end{tabular}
\end{table}

\textbf{CTAB-GAN}. To understand whether our feature sorting algorithm works when sparsity is involved, we test it on CTAB-GAN, and the results are summarized in Table \ref{tab: ctabgan-feature-sorting}. Surprisingly, the algorithm can reduce Dif. Corr. and WD by more than 10 \% on all datasets except the Credit dataset. On the Loan dataset, the Dif. Corr. and WD are decreased by 57\% and 29\% after feature sorting, meaning that the algorithm can effectively improve the statistical similarity between synthetic and real data. 


Compared with table-GAN, CTAB-GAN has more performance gain after feature sorting. This is due to the sparsity issue caused by the encoding methods of CTAB-GAN, i.e., mode-specific normalization for numerical features and one-hot encoding for categorical features. Since the input data are sparse after encoding, putting the highly-correlated columns together can drastically reduce the distance between correlated columns, and therefore improves CTAB-GAN's ability to capture the relation between highly-correlated columns. 

To summarize, our feature sorting algorithm can improve the performance of CNN-based table synthesizers, especially when the input tabular data are sparse. For dense tabular data, it also works if the relation between correlated features is relatively simple.

\begin{table}[t]
\centering
\caption{ CTAB-GAN before and after the feature sorting algorithm.
}
\label{tab: ctabgan-feature-sorting}
\begin{tabular}{@{}ccccccc@{}}
\toprule
\multirow{2}{*}{Dataset} & \multicolumn{2}{c}{Before sorting} & \multicolumn{2}{c}{After sorting} & \multicolumn{2}{c}{Change} \\ 
\cmidrule(l{4pt}r{4pt}){2-3}
\cmidrule(l{4pt}r{4pt}){4-5}
\cmidrule(l{4pt}r{4pt}){6-7}
                                  & {Dif. Corr.}      & {WD}      & {Dif. Corr.}      & {WD}     & {Dif. Corr.}  & {WD}  \\ \midrule
Loan                              & 1.469                    & 0.356            & 0.638                    & 0.253           & -57\%                & -29\%        \\
Adult                             & 0.448                    & 1.517            & 0.296                    & 1.205           & -34\%                & -21\%        \\
Credit                            & 1.688                    & 0.115            & 1.660                    & 0.134           & -2\%                 & 17\%         \\
Covtype                           & 1.948                    & 0.539            & 1.442                    & 0.475           & -26\%                & -12\%        \\
Intrusion                         & 3.969                    & 2.668            & 3.385                    & 1.999           & -15\%                & -25\%        \\ \midrule
Avg.                              & 1.904                    & 1.039            & 1.484                    & 0.813           & -22\%                & -22\%        \\ \bottomrule
\end{tabular}
\end{table}

\section{Conclusion}
Motivated by the soaring need of synthetic big data, we discover and analyze the varying performance of AI-based tabular data synthesizers to the input column permutation. The state-of-the-art tabular data synthesizers, especially the ones based on convolution neural networks, are sensitive to the column order of the training input. Through empirical analysis on extensive combinations of column permutations, synthesizers, and datasets, we find the root causes for lacking column permutation invariance are the data representation of tabular data and use of convolution neural networks. To address these limitations, we first propose AE-GAN, a GAN-based synthesizer leveraging the representation capacity of autoencoder. Secondly, we propose a feature sorting algorithm that preserves the correlation across input columns and enhances the-state-of-the-art synthesizers considered. Our evaluation results on five datasets show that AE-GAN makes the excellent trade-off among the sensitivity to the input column permutation, training time and the high synthetic data quality and utility. The proposed feature sorting algorithm, on the other hand, enhances the synthesis quality of exiting CNN-based synthesizers, i.e., the statistical difference and ML utility difference between real and synthetic data, by 22\%. 


\bibliographystyle{ieeetr}
\bibliography{Ref} 

\end{document}